\documentclass{sig-alternate}
\usepackage{todonotes}
\usepackage{enumitem}
\usepackage{url}
\usepackage{algorithm}
\usepackage{algorithmicx}
\usepackage{algpseudocode}
\usepackage{amssymb}
\usepackage{amsmath}
\usepackage{mathtools}

\setdescription{leftmargin=\parindent,labelindent=0pt}

\begin{document}
\sloppy
\conferenceinfo{GECCO'13,} {July 6-10, 2013, Amsterdam, The Netherlands.}
    \CopyrightYear{2013}
    \crdata{TBA}
    \clubpenalty=10000
    \widowpenalty = 10000

\title{Generic Behaviour Similarity Measures\\for Evolutionary Swarm Robotics}

%
%
%
%
%

\numberofauthors{2}
\author{
\alignauthor
Jorge Gomes\\
       \affaddr{Instituto de Telecomunica\c{c}\~oes} \& LabMAg-FCUL\\
       \affaddr{Lisbon, Portugal}\\
       \email{jgomes@di.fc.ul.pt}
\alignauthor
Anders L. Christensen\\
       \affaddr{Instituto de Telecomunica\c{c}\~oes \& ISCTE-IUL}\\
       \affaddr{Lisbon, Portugal}\\
       \email{anders.christensen@iscte.pt}
}


\date{31 January 2013}

\maketitle
\begin{abstract}
Novelty search has shown to be a promising approach for the evolution of controllers for swarm robotics. In existing studies, however, the experimenter had to craft a domain dependent behaviour similarity measure to use novelty search in swarm robotics applications. The reliance on hand-crafted similarity measures places an additional burden to the experimenter and introduces a bias in the evolutionary process. In this paper, we propose and compare two task-independent, generic behaviour similarity measures: \emph{combined state count} and \emph{sampled average state}. The proposed measures use the values of sensors and effectors recorded for each individual robot of the swarm. The characterisation of the group-level behaviour is then obtained by combining the sensor-effector values from all the robots. We evaluate the proposed measures in an aggregation task and in a resource sharing task. We show that the generic measures match the performance of domain dependent measures in terms of solution quality. Our results indicate that the proposed generic measures operate as effective behaviour similarity measures, and that it is possible to leverage the benefits of novelty search without having to craft domain specific similarity measures.
\end{abstract}

\category{I.2}{Artificial Intelligence}{Robotics}
\category{I.2}{Artificial Intelligence}{Distributed Artificial Intelligence}

\terms{Algorithms}

\keywords{Evolutionary swarm robotics, novelty search, behaviour characterisation} 

\section{Introduction}

Swarm robotics is a promising approach to collective robotics, where the group level behaviour emerges from the local interactions among agents, and from the interactions between the agents and the environment~\cite{sahin05a}. This approach has the potential to incite several desirable properties in a group of agents, such as robustness, flexibility, and scalability~\cite{sahin05a}. However, the complexity stemming from the intricate dynamics required to produce self-organised behaviour complicates the hand-design of control systems~\cite{trianni03}. Artificial evolution has been shown capable of exploiting the intricate dynamics and synthesise self-organised behaviours (see for example~\cite{trianni03, trianni06a, baldassarre03, baldassarre07}), but the approach carries several issues. The most prominent issue associated with common evolutionary techniques is \emph{deception}~\cite{whitley91}. Deception occurs when the fitness function misguides the search towards local maxima that do not contain adequate solutions to the problem. As the complexity of a problem increases, the fitness landscape typically becomes more rugged and gains more local maxima~\cite{nelson09}. As such, it becomes more difficult to craft a fitness function that can successfully guide the search towards the objective~\cite{zaera96}, i.e., the evolutionary process becomes more vulnerable to deception. 

Novelty search~\cite{stan11a} is a distinctive evolutionary approach where candidate solutions are rewarded based solely on their behavioural novelty, with respect to previously evaluated solutions. In recent work~\cite{gomes12a,gomes12b}, it was shown that novelty search can avoid deception in the evolution of swarm robotic systems. Besides not being affected by deception, it was also shown that novelty search is able to find a greater diversity of solutions, and the successful solutions were simpler in terms of neural network complexity, when compared to those found by fitness-based evolution. But these advantages come at a price: for novelty search to work, it is necessary to craft the domain dependent behaviour similarity measure, used to compute the novelty of the individuals. The results showed that the choice of the novelty metric has a significant impact in the performance of novelty search, and can introduce a significant bias in the evolutionary process.

Previous works have proposed behaviour similarity measures that are domain independent~\cite{gomez09, doncieux10, mouret12}. These generic measures can potentially be used to overcome the aforementioned limitation of novelty search. Generic measures are typically based on the sensor and effector values of the agents exclusively, and do not rely on domain knowledge provided by the experimenter. However, the generic measures described in previous works are aimed at single robotic systems. In this paper, we study how generic measures can be adapted to swarm robotic systems.

This paper proposes generic behaviour similarity measures that use the sensor and effector values of the robots of the swarm to obtain a representation of the typical behaviour of the swarm as whole. The measures are evaluated in two swarm robotics tasks: (i) an aggregation task; and (ii) a task where robots must share an energy recharging station in order to survive. Following previous results~\cite{gomes12b}, novelty search is used in combination with the fitness function, through a linear scalarization. of novelty and fitness objectives. NEAT is used as the underlying neuroevolution method.

The results of our experiments suggest that novelty search with the proposed generic similarity measures can match the performance of novelty search with domain-dependent similarity measures, regarding the quality of the evolved solutions. We show that the documented advantages of novelty search, such as its capacity to bootstrap evolution and to circumvent deception~\cite{gomes12a}, are also present with the use of generic measures.


\section{Related Work}

In this section, we describe the novelty search algorithm, and how novelty search can be combined with fitness-based evolution to improve the effectiveness of the evolutionary process. We move on to discuss the previous applicatons of novelty search in evolutionary robotics. We conclude with a discussion of the generic behaviour similarity measures proposed in previous works.

\subsection{Novelty Search}

Novelty search~\cite{stan11a} can be implemented over any evolutionary algorithm. The distinctive aspect of novelty search is how the individuals of the population are scored. Instead of being scored according to how well they perform a given task -- which is typically measured by a fitness function, the individuals are scored based on their behavioural novelty -- which is given by the novelty metric. This metric quantifies how different an individual is from the other, previously evaluated individuals with respect to behaviour. 

To measure how far an individual is from others individuals in behaviour space, the novelty metric relies on the average distance of that individual to the $k$-nearest neighbours, among the current population and a sample of the previously seen behaviours (stored in an archive). The behaviour distance between each two individuals is given by a function $dist$ that should be provided by the experimenter. Candidates from sparse regions of the behaviour space thus tend to receive higher novelty scores, thereby creating a constant evolutionary pressure towards behavioural innovation. 

The function $dist$ is typically defined with domain knowledge. Following this approach, the behaviour of each individual is characterised by a vector of real numbers. The experimenter should design the behaviour characterisation vector so that it captures behaviour features that are considered relevant to the problem or task. The behaviour distance between two individuals is then given as the Euclidian distance between the corresponding behaviour characterisation vectors of the individuals. A distinct approach is to use distance functions that do not rely on domain knowledge. This approach is the main focus of this paper and will be detailed in Section~\ref{sec:generic_measures}.

\subsubsection{Combining Novelty and Fitness}
\label{sec:combination}

As novelty search is guided by behavioural innovation alone, its performance can be greatly affected by the size and shape of the behaviour space. In particular, behaviour spaces that are vast or contain dimensions not related with the task can negatively impact the performance of novelty search~\cite{stan10a,cuccu11b}, because novelty search may spend most of its time exploring behaviours that are irrelevant for the goal task. To address this issue, several authors have proposed techniques that combine novelty with fitness in the evaluation of the individuals~\cite{stan10a,cuccu11b,gomes12b,mouret11,mouret12}.

In our experiments, we use a linear scalarization of the novelty and fitness objectives~\cite{cuccu11b}. We chose this approach because it can be used together with NEAT without any further modifications, and has shown promising results in previous studies~\cite{gomes12b}. Linear scalarization of the novelty and fitness objectives directs the search towards regions with high fitness in the behaviour space. An individual $i$ is evaluated to measure both fitness, $f\!it(i)$, and novelty, $nov(i)$, which after being normalised (Eq.~\ref{eq:blendnorm}) are combined according to Eq.~\ref{eq:blend}.
\begin{equation}
\overline{f\!it}(i)=\frac{f\!it(i)-f\!it_{min}}{f\!it_{max}-f\!it_{min}} \; , \;  \overline{nov}(i)=\frac{nov(i)-nov_{min}}{nov_{max}-nov_{min}}
\label{eq:blendnorm}
\end{equation}
\begin{equation}
score(i)=(1-\rho)\cdot \overline{f\!it}(i)+\rho \cdot \overline{nov}(i)
\label{eq:blend}
\end{equation}

The parameter $\rho$ controls the relative weight of novelty, and must be specified by the experimenter. $f\!it_{min}$ and $nov_{min}$ are the lowest fitness and lowest novelty score in the current population, and $f\!it_{max}$ and $nov_{max}$ are the highest fitness and highest novelty score, respectively.

\subsubsection{Novelty Search in Evolutionary Robotics}

Novelty search, and other evolutionary techniques based on behavioural diversity, have been applied with success to single robotic systems. Some of these applications include body-brain co-evolution~\cite{krcah10}; biped robot control~\cite{stan11a}; robot navigation in deceptive mazes~\cite{stan11a}; sequential light seeking~\cite{mouret12}; and a robot ball-collecting task~\cite{mouret12}. In~\cite{mouret12}, it is presented a comprehensive study of the use of diversity-based techniques in evolutionary robotics.

Gomes et al.~\cite{gomes12a,gomes12b} showed that novelty search can also provide a valuable contribution to the evolution of controllers for swarm robotics. In particular, the results showed that the use of novelty search circumvented deception and bootstrapping problems, and could unveil a broad diversity of solutions for the same problem. However, the same studies revealed that defining behaviour characterisations for this domain can be a delicate endeavour. Since there are infinitely many behavioural possibilities, many of these possibilities must be conflated in order to construct a viable search space. Excessive conflation, however, can hinder the evolution of certain types of solutions, and degrade the performance of novelty search. Furthermore, the definition of the behaviour characterisation adds a human bias to the process, which is an aspect that should be minimised in evolutionary robotics~\cite{nelson09}.

\subsection{Generic Novelty Measures}
\label{sec:generic_measures}

Gomez~\cite{gomez09} proposes the use of generic measures for assessing the behaviour similarity between individuals. The proposed approach consists of building state-action trajectories for the agent, i.e., the history of actions of the agent through time. These trajectories are then compared, obtaining a measure of behaviour similarity, without the need of providing domain-specific knowledge. To compare the sequences of actions, the author evaluates the use of Hamming distance, relative entropy, and normalised compression distance (NCD). The experimental setup is based on the Tartarus problem, and the results show that the NCD distance offers the best performance, followed closely by the Hamming distance. NCD is a similarity measure that exploits the algorithmic regularities of the sequences, but introduces a significant computational overhead.

To address the difficulty in designing behaviour characterisations for evolutionary robotics, Doncieux and Mouret~\cite{doncieux10} proposed and compared generic behaviour similarity measures for evolutionary robotics. Any evolutionary robotics experiment involves robots with actuators and sensors, whose values reflect the microscopic behaviour of the robot. This notion led to the definition of the following generic measures~\cite{doncieux10}:

\begin{description}
\item[Hamming distance] A vector is built with the sensor and effector values of the robot, sampled throughout the simulation:
\begin{equation}
\vartheta = [\{\mathbf{s}(t),\mathbf{e}(t)\},t\in [0,T]] \enspace ,
\end{equation}
where $\mathbf{s}(t)$ and $\mathbf{e}(t)$ are the vectors of the sensor and effector values at time $t$, respectively, and $T$ is the simulation time. The vector $\vartheta$ is then binarised into $\vartheta_{bin}$, by transforming each value in either 0 or 1. The similarity measure is then given as the Hamming distance between the corresponding $\vartheta_{bin}$ vectors obtained for each individual.

\item[Direct Fourier Transform]
The $\vartheta$ vectors are obtained for each individual, similar to the Hamming distance measure. But instead of using the complete vectors, a Discrete Fourier Transform (DFT) is used to reduce the dimensionality. The similarity measure is defined as the Euclidean distance between the first $n_{F}$ coefficients of the DFT.

\item[Systematic State Count]
Perception-action states are defined based on the possible combinations of $\vartheta_{bin}$. Relying on the sensor-effector data, the number of times the robot was in a particular state is then evaluated, resulting in a vector of $n$ integers, $n$ being the number of such states. The similarity measure is then defined as the mean element-wise distance between the vectors.
\end{description}

These methods were evaluated in a ball collection task, where the robot had 9 sensors and 3 effectors. The novelty metric was combined with the fitness function through multi-objectivisation. The results showed that the Hamming distance measure was the most effective, being superior to the similarity measure defined with domain knowledge. The systematic state count and DFT measures displayed a significantly lower performance when compared to the Hamming distance.

The Hamming distance similarity measure was further tested in~\cite{mouret12}. In these experiments, the measure was evaluated in three different tasks (deceptive maze, sequential light seeking, ball-collecting robot), and different diversity maintenance techniques. When using multi-objectivisation of novelty and fitness, the results showed that the generic Hamming distance was at least as good as the similarity measure manually defined with domain knowledge, regarding the quality of the evolved solutions.

\newpage

\section{Methods}

\subsection{Combined State Count}


The proposed \emph{Combined State Count} is an adaptation of \emph{Systematic State Count} (see Section~\ref{sec:generic_measures}). Despite the lower performance in the experiments of Doncieux \& Mouret~\cite{doncieux10}, when compared to the other generic measures, the concept of this method can be directly adapted to swarms of robots. As such, it is the starting point of our study. The principle is to define states based on the values from the sensors and effectors recorded for each robot. Then, the number of times the robots of the swarm were in each state is computed. There is no discrimination in terms of which robot was in a particular state, i.e, the state counting at the swarm level is the sum of the state counts for each robot in the swarm.

The state counting approach is, however, prone to suffer from scalability issues, since the number of states grows exponentially with the number of sensors/effectors, and with the number of possible values for each sensor/effector. To address this issue, we propose modifications over the original \emph{State count}. Scalability is achieved through the use of efficient structures for representing states and characterisations, and mechanisms for reducing the effective number of states.

\subsubsection*{Efficient State Count Representation}

Representing each behaviour characterisation as a vector with one position for each state (as proposed in~\cite{doncieux10}) can compromise the efficiency of the algorithm if there is a large number of states. However, the number of visited states in one simulation is only a small fraction of the total number of possible states. As such, we can represent each characterisation as a map from states to counts. The counting is normalised according to the size of the swarm, to allow fair comparisons between simulations with different swarm sizes. The behaviour similarity measure is then given by the difference between the state count maps. To calculate the characterisation map $m'$ for each individual, Algorithm~\ref{alg:statecount} is used:

\begin{algorithm}[H]
\begin{algorithmic}
\State $m \gets Map<Int, Float>$ 
\ForAll{simulation-steps}
\ForAll{$r$ in robots}
\State $\vartheta_{r}  \gets$ read-state($r$)
\State $\vartheta'_{r} \gets$ discretise($\vartheta_{r}$)
\State $h \gets$ hash($\vartheta'_{r}$)
\If{$m$ does not contain $h$}
	\State $m[h] \gets 0$
\EndIf
\State $m[h] \gets m[h] + 1/swarmsize$
\EndFor
\EndFor
\State $m' \gets$ filter($m$)
\State \Return $m'$
\end{algorithmic}
\caption{State count characterisation}
\label{alg:statecount}
\end{algorithm}

The function \emph{read-state} retrieves the current sensor-effector state $\vartheta(r)$ for a particular robot $r$: 
\begin{equation}
\vartheta(r) = \{\mathbf{s}(r),\mathbf{e}(r)\} \enspace ,
\end{equation}
where $\mathbf{s}(r)$ is the vector of size $n_{s}$, composed of the values coming from the $n_{s}$ sensors of the robot $r$; and $\mathbf{e}(r)$ is the vector composed of the effector values.

The discretised vector $\vartheta'(r)$ is obtained by independently normalising each element of $\vartheta(r)$ to the interval $[0,K-1]$, followed by an approximation to the nearest integer:
\begin{equation}
\vartheta'_{i}(r) = \left \| \frac{\vartheta_{i}(r)-\vartheta_{i,max}}{\vartheta_{i,max}-\vartheta_{i,min}} \cdot (K-1)\right \| \enspace ,
\end{equation}
where $\vartheta_{i,max}$ and $\vartheta_{i,min}$ are respectively the maximum and minimum values of the $i$-th sensor/effector, and $K$ is the number of target partitions. The parameter $K$ has direct implications in the number of possible states, and it should be empirically determined. A rule of thumb is to define it accordingly to the length of $\vartheta$. For most applications, $K$ values of 2 and 3 are adequate, categorising the value of each sensor in High/Low, or High/Medium/Low. However, if the robots have a small number of sensors ($\vartheta$ is relatively short), higher values of $K$ might be preferred, in order to operate with more detailed behaviour characterisations.

The function \emph{hash} was implemented with the Jenkins' one-at-a-time hash.\footnote{\url{http://www.burtleburtle.net/bob/hash/doobs.html}} The intent of hashing the vector $\vartheta'(r)$ is twofold. First, it allows lookups of the corresponding entry in the $m$ map in $\mathcal{O}(n)$ time, $n$ being the length of $\vartheta'(r)$. Second, as different vectors are hashed to different values (there is a very low probability of collisions), there is no need to store $\vartheta'$ vectors, which improves the space complexity of the algorithm.

\subsubsection*{Reducing the Number of States}

The function \emph{filter} eliminates the least observed states, in order to improve the efficiency of the algorithm. Preliminary results revealed that robots tend to spend most of their time in a small subset of the state space. Most of the states are visited only in one or a few simulation steps. As such, eliminating these states from the behaviour characterisation can significantly improve the efficiency of the algorithm, practically without compromising the accuracy of the characterisation. The function \emph{filter} removes from the characterisation the states where the robots spent less than $T\%$ of the time: 
\begin{equation}
m' = \left \{ (h,c) \in m : c > \sum_{i \in m} m[i] \cdot T  \right \} \enspace .
\end{equation}

The constant $T$ should be empirically determined. In our experiments, a value of only 1\% was enough to drastically reduce the number of states in each characterisation. For instance, the preliminary results showed that on average 99\% of the simulation time was spent on only 10\% of the visited states. 

\subsubsection*{Distance Between Characterisations}

To calculate the distance between two characterisations we chose to use the Bray-Curtis dissimilarity, a well-known measurement for quantifying the difference between samples of abundance data. Bray-Curtis is a modified Manhattan measure, where the summed differences between the variables are standardised by the summed variables of the samples. This measure is within the range of 0 to 1. A value of 0 indicates that the two samples have the same composition, while a value of 1 means the two samples do not share any element. 

Adapting the Bray-Curtis dissimilarity behaviour characterisations, the difference $b$ between two characterisations $m_{1}$ and $m_{2}$ is given by:
\begin{equation}
\begin{multlined}
b(m_{1}, m_{2}) = \\
\frac{\displaystyle\sum_{i \, \in \, m_{1}\cap m_{2}} \hspace{-7pt} \left | m_{1}[i]-m_{2}[i] \right | + \hspace{-7pt} \sum_{i \, \in \, m_{1}\setminus m_{2}} \hspace{-7pt} m_{1}[i] + \hspace{-7pt} \sum_{i \, \in \, m_{2}\setminus m_{1}} \hspace{-7pt} m_{2}[i]}
{\displaystyle\sum_{i \, \in m_{1}} m_{1}[i] + \sum_{i \, \in m_{2}} m_{2}[i]} \enspace .
\end{multlined}
\end{equation}


\subsection{Sampled Average State}


The second similarity measure relies on the principles of the Hamming Distance measure (see Section~\ref{sec:generic_measures}), which was one of the most successful generic similarity measures in previous works with single robotic systems~\cite{doncieux10,mouret12}. However, this measure relies on the full description of the sensor-effector states of the robot through time.  As such, it can not be directly used with swarms of robots because (i) it would not scale with the number of robots, and (ii) the behaviour of an individual robot in a swarm often has a significant stochastic component. To overcome these issues, we propose the following modifications:


\begin{itemize}
\item The state of the swarm at a given instant is the average of the sensor-effector states of each robot. This allows scalability in respect to the size of the swarm.
\item The state of the swarm is averaged over a certain time window. This reduces the sensitivity to the initial conditions, and to the stochastic nature of the individual robots behaviour. 
\end{itemize} 


The characterisation of an individual is given by:
\begin{equation}
\vartheta = \left [\{ \overline{v_{0}}(w), \cdots, \overline{v_{n}}(w) \}, w \in [0,W[ \; \right ] \enspace ,
\end{equation}
where $W$ is the number of time windows and $\overline{v_{i}}(p)$ is the average value of the $i$-th sensor/effector over the $w$-th time window:
\begin{equation}
\overline{v_{i}}(w) = \frac{W}{T}\sum_{t=wT/P}^{(w+1)T/P} \frac{1}{R} \sum_{r=0}^{R-1}v'_{i,r}(t) \enspace ,
\end{equation}
$T$ is the total simulation time, $R$ the number of robots, and $v'_{i,r}(t)$ is the normalised value of the $i$-th sensor/effector of the robot $r$, at instant $t$:
\begin{equation}
v'_{i,r}(t) = \frac{v_{i,r}-v_{i,max}}{v_{i,max}-v_{i,min}} \enspace ,
\end{equation}
$v_{i,max}$ and $v_{i,min}$ are the maximum and minimum values of the $i$-th sensor/effector, respectively.

The distance between two characterisations $\vartheta_{1}$ and $\vartheta_{2}$ is then given by the Manhattan distance between the vectors:
\begin{equation}
d_{man}(\vartheta_{1},\vartheta_{2})= \sum |\vartheta_{1}[i]-\vartheta_{2}[i]| \enspace .
\end{equation}


\section{Experimental Setup}

The proposed generic similarity measures are evaluated over two swarm robotic tasks: aggregation and resource sharing. The generic measures are compared with domain dependent measures, and with fitness-based evolution.

Our experimental framework is based on Simbad 3d Robot Simulator~\cite{hugues06} for the robotic simulations. In both tasks, the environment is a 3\,m by 3\,m square arena bounded by walls. The swarms are homogeneous. Each robot is modelled after the e-puck, but with modifications to the sensor setup. Each robot is circular with a diameter of 8\,cm, and is equipped with differential drive, capable of delivering speeds of up to 12\,cm/s. The local on-board controllers are recurrent neural networks. The inputs of the neural networks are the normalised values of the sensors of the robot, and there are three outputs: one to control each of the two motors, and one dedicated to completely halt the movement of the robot. Each simulation lasts for 2500 simulation steps, which corresponds to 250\,s of simulated time.

\subsection{Aggregation Task}

Aggregation is a commonly studied task in swarm robotics~\cite{trianni03,baldassarre03}. In this task, a dispersed robot swarm must form a single cluster in any point of the arena. The swarm has a fixed size of 7 robots. Each robot is equipped with (i) 8 IR sensors evenly distributed around its chassis for the detection of obstacles (walls or other robots) within a range of 10\,cm; (ii) 8 IR sensors dedicated to the detection of other robots within a range of 25\,cm; and (iii) a sensor that returns the percentage of nearby robots (within a radius of 25\,cm), relative to the swarm size.

The fitness function $F_{a}$ is defined as the average distance of the robots to the centre of mass of the swarm, measured at the last instant of the simulation:
\begin{equation}
F_{a}=1-\sum_{i=1}^{N}\frac{dist(\mathbf{R}_T,\mathbf{r}_{i_T})}{N} \enspace ,
\end{equation}
where $\mathbf{R}_T$ is the centre of mass in the last instant of simulation, and $\mathbf{r}_{i_{T}}$ is the position of robot $i$. The distance values are normalised to $[0,1]$.

The domain dependent behaviour characterisation, used as benchmark, is based on the average distance to the centre of mass of the swarm, and the number of clusters, sampled through the simulation time~\cite{gomes12a}. Considering a simulation with $N$ robots and $T$ temporal samples, the characterisation $\mathbf{b_{a}}$ is given by:
\begin{align}
\mathbf{b_{a}}&=\{\mathbf{cm},\mathbf{cl}\} \notag \\
\mathbf{cm}&=\frac{1}{N} \left [ \sum_{i=1}^{N}d(\mathbf{R}_1,\mathbf{r}_{i_1}),\cdots , \sum_{i=1}^{N}d(\mathbf{R}_T,\mathbf{r}_{i_T}) \right ] \\
\mathbf{cl}&=\frac{1}{N}\left [clusterCount(1), \cdots , clusterCount(T) \right ] \enspace ,\notag
\end{align}
where $\mathbf{R}_t$ is the centre of mass at time $t$, and $\mathbf{r}_{i_{t}}$ is the position of robot $i$ at time $t$. The function $d$ gives the distance normalised in the range $[0,1]$. The function $clusterCount$ returns the number of robot clusters. Two robots belong to the same cluster if the distance between them is less than the robot IR sensor range (25\,cm).

\subsection{Resource Sharing Task}

In this task, the swarm must coordinate in order to allow each member periodical access to a single battery charging station. The robots should first find the charging station, and then effectively share the station to ensure the survival of all the robots in the swarm. The charging station can only hold one robot at the time.

Our experiments use a group of 3 robots. Each robot has (i)~8 IR sensors for the detection of obstacles up to a range of 10\,cm; (ii)~8 sensors dedicated to the detection of other robots up to a range of 25\,cm; (iii)~8 sensors for the detection of the charging station up to a range of 1\,m; (iv)~a binary sensor that indicates if the robot is over the charging station; and (v)~a proprioceptive sensor that reads the current energy level of the robot. 

Each robot starts with full energy (1000 units), and spends energy at a rate proportional to motor usage: a robot spends 5 units per second when motors are off, and 10 units of energy per second when motors propel the robot at its maximum speed. The charging station is placed in the centre of the arena, and charges a robot at a rate of 100 units of energy per second. The robots have to be completely stopped in order to charge.

The fitness function $F_{s}$ used to evaluate the controllers is a linear combination of the number of robots alive at the end of the simulation and the average energy of the robots throughout the entire simulation:
\begin{equation}
F_{s} = 0.9\cdot\frac{|a_{T}|}{N}+0.1\cdot \sum_{t=1}^{T}\sum_{i=1}^{N}\frac{e_{i_{t}}}{TNe_{max}}\enspace ,
\end{equation}
where $|a_{T}|$ is the number of robots still alive in the end of the simulation, $T$ is the length of the simulation, $N$ is the number of robots in the swarm, $e_{i_{t}}$ is the energy of the robot $i$ at time $t$, and $e_{max}$ is the maximum energy of a robot. The second term of $F_s$ concerning the average energy is included to differentiate solutions where the same number of robots survive.

The domain dependent behaviour characterisation is an extension of the characterisation used in previous experiments with this task~\cite{gomes12b}. The characterisation is a vector of length four, composed by the following behavioural features that are related to the task: (i) The number of robots that reached the end of the simulation alive; (ii) the average energy of the alive robots throughout the simulation; (iii) the average movement of all alive robots; and (iv) the average distance of all alive robots to the charging station. Each of these elements is normalised to $[0,1]$.

\subsection{Configuration of the Algorithms}
\label{sec:treatments}

NEAT~\cite{stan02} is used as the underlying neuroevolution algorithm. NEAT is widely used, and one of the most successful neuroevolution approaches developed to date. We use the implementation provided in NEAT4J.\footnote{NeuroEvolution for Augmenting Topologies for Java -- \url{http://neat4j.sourceforge.net}} The parameters for NEAT were the same in all experiments: recurrent links are allowed, crossover rate -- 25\%, mutation rate -- 10\%, population size -- 200. The remaining parameters were assigned their default value in the NEAT4J implementation.

The implementation of novelty search follows the description in~\cite{stan11a}. We used a $k$ value of 15 nearest neighbours, and the individuals are added to the novelty archive with a probability of 2\%~\cite{stan10b}. The size of the archive is bounded to 500 individuals. When the archive is full, individuals are randomly removed as needed.

Novelty search is combined with the fitness-function through a linear scalarization. of the novelty and fitness objectives (see Section~\ref{sec:combination}). In all novelty search experiments, the value of $\rho$ was set to 0.7, which means that the score of each individual is based on 70\% of the novelty score and 30\% of the fitness score. This value was empirically chosen, and in agreement with previous experiments~\cite{gomes12b}.

For the \emph{combined state count} measure, the filter threshold $T$ was set to 1\% in all experiments, and the discretisation level $K$ was set to 3. For the \emph{sampled average state} measure, three values of $W$ were tested: 1, 10, and 50, which correspond to time windows of 250\,s, 25\,s, and 5\,s, respectively. In both generic similarity measures, the values coming from the sensor arrays (composed of 8 sensors for the detection of obstacles, other robots, or the charging station) were compressed in four values. These four values represent the closest distance measured at the front of the robot, left, right, and back. This compression was done to reduce the number of states (in the \emph{combined state count} measure), and to reduce the length of the characterisation (in the \emph{sampled average state} measure). 

Each controller was evaluated in 10 simulations, randomly varying the initial positions and orientations of the robots. The fitness scores obtained in each simulation are combined to a single value using the harmonic mean as advocated in~\cite{sahin05b}. The behaviour characterisations obtained in the multiple simulations are also merged in a single one through an element-wise average (in the domain dependent measures and \emph{sampled average state}), and by summing the state counts (in \emph{combined state count}). The best individuals found in each generation were post-evaluated with 50 simulations, in order to attain more reliable statistics.


\section{Results}

The following treatments were applied to each task. Each evolutionary method was evaluated in 10 independent evolutionary runs. The parameters of each method were set as specified in Section~\ref{sec:treatments}.
\begin{description}[noitemsep,labelwidth=1cm]
\item[SC] Combined state count
\item[AS-1] Sampled average state with $W=1$
\item[AS-10] Sampled average state with $W=10$
\item[AS-50] Sampled average state with $W=50$
\item[DD] Novelty with domain dependent similarity measure
\item[Fit] Fitness-based evolution
\end{description}

\begin{figure}[ht!]
	\centering
	\includegraphics[width=1\columnwidth]{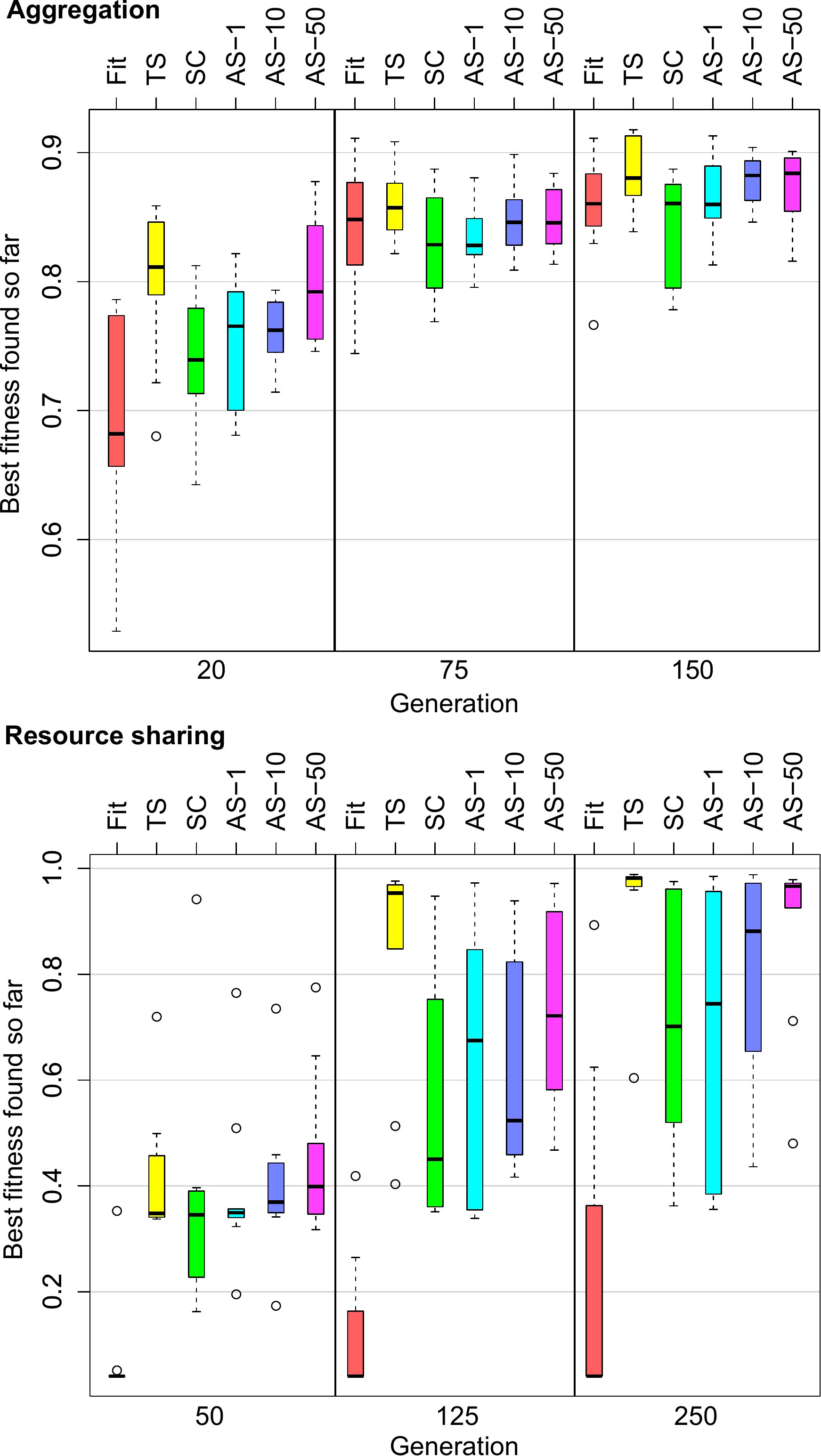}
	\caption{Performance comparison of the evolutionary treatments in both tasks, regarding the highest fitness score achieved at different stages of the evolutionary process. The boxplots represent the distribution of the fitness scores obtained in the 10 evolutionary runs of each treatment.}
	\label{fig:all_boxplots}
\end{figure}

The quality of the solutions evolved with each evolutionary method is depicted in Figure~\ref{fig:all_boxplots}. The boxplots represent the highest fitness score found until a given generation, in each evolutionary run of each treatment. The depicted results are further explained next.

\subsection{Aggregation}



As the results show (Figure~\ref{fig:all_boxplots} -- Aggregation), the fitness function is not deceptive, as fitness-based evolution can almost always reach good quality solutions. The most notorious advantage of novelty search is its capacity of avoiding deception. However, previous work~\cite{gomes12a} has shown that even in non-deceptive swarm robotic tasks, novelty search can offer a number of advantages. As such, it is still valuable to analyse the performance of novelty search with generic behaviour similarity measures in this non-deceptive task.

In early stages of evolution (at generation 20), novelty search has an advantage over fitness-based evolution, confirming that novelty search quickly bootstraps the evolutionary process~\cite{gomes12a,mouret09}. All similarity measures, except for \emph{state count} were superior to fitness-based evolution ($p$-value $<$ 0.05, Mann-Whitney U test).

Around the middle of the evolution (generation 75), the differences between the multiple treatments are less pronounced. By the end of the evolution, the domain dependent similarity measure is only superior to the state count measure ($p$-value $<$ 0.05). This absence of significant difference between treatments is actually a promising result. Previous work~\cite{gomes12a} has shown that when the behaviour similarity measure is poorly defined, the performance of novelty search tends to degrade significantly, regarding the quality of the solutions evolved. In our experiment, the generic measures yielded results similar to fitness-based evolution and to the domain dependent measure, which suggests that the generic measures are indeed acting as effective behaviour similarity measures.

\subsection{Resource Sharing}



As previous experiments have shown~\cite{gomes12b}, the resource sharing task is inherently deceptive. In particular, fitness-based evolution tends to get stuck in two local maxima: (i) The robots do not move at all in order to conserve energy and survive longer, and as a consequence, they can not find the charging station and all the robots run out of energy (fitness score around 0.04); and (ii) when a robot finds the charging station, it occupies it and never leaves, condemning the other robots (fitness score around 0.38). The deceptiveness of this task makes it especially suitable to solve using novelty search. As such, this task is a good benchmark to evaluate if the behaviour similarity measures are capable of avoiding deception and guiding evolution towards good solutions.

At the early stages of evolution (generation 50, see Figure~\ref{fig:all_boxplots} -- Resource sharing), almost all runs of fitness-based evolution are still stuck in the local maximum where the robots do not move. On the other hand, all treatments based on behaviour novelty could successfully bootstrap the evolution. At this early stage, there are still no significant differences between the novelty based treatments. By the middle of the evolutionary process, the domain dependent similarity measure stands out, being superior to all the treatments ($p$-value $<$ 0.05, Mann-Whitney U test), except for \textbf{AS-50}. There are no statistically significant differences between generic similarity measures at this stage.

Looking at the best fitness scores achieved in the whole evolution (generation 250), the superiority of the domain dependent similarity measure holds. However, all novelty based treatments were superior to fitness based-evolution ($p$-value $<$ 0.05), and more or less consistently, all reached high fitness scores. Regarding the generic similarity measures, the \textbf{AS-50} treatment stands out, being significantly superior to \textbf{SC} and \textbf{AS-1} ($p$-value $<$ 0.05).

\subsection{Combined State Count}

\begin{figure}[!b]
	\centering
	\includegraphics[width=1\columnwidth]{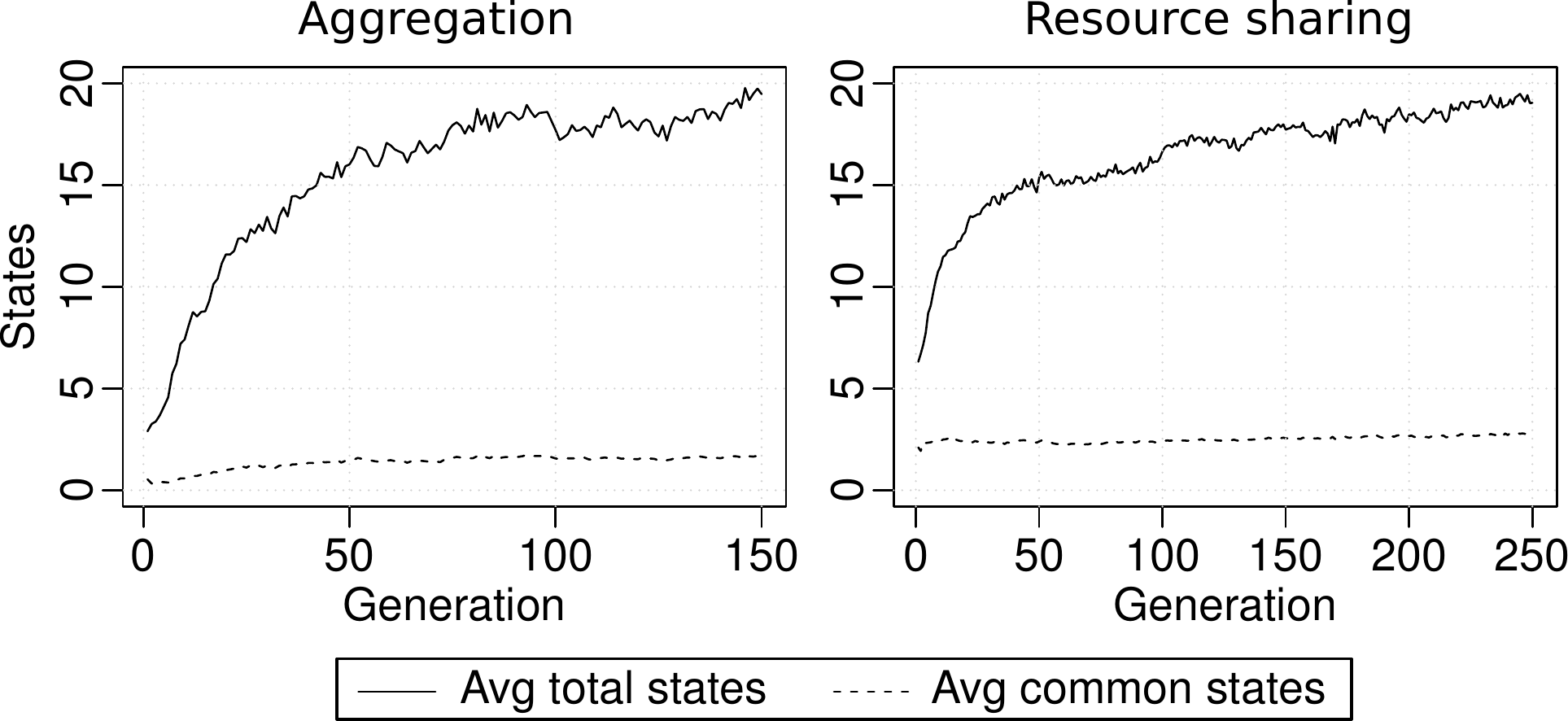}
	\caption{Average number of sensor-effector states visited by each population individual (after the filtering step), compared with the average number of states that each individual shares with the current population and the novelty archive.}
	\label{fig:statecount}
\end{figure}

In both tasks, the \emph{combined state count} measure was the least effective generic measure. Nevertheless, the performance was close to the \emph{sampled average state}, which contrasts with the results in~\cite{doncieux10}. To shed some light on the inferior performance of \emph{combined state count}, we analysed the sensor-effector states that are visited with each individual (Figure~\ref{fig:statecount}). 

The increasing average number of states depicts the increasing complexity of the solutions, throughout the evolution. However, the average number of common states do not follow this trend. Since the distance between two state count characterisations is essentially determined by the states they share, this distance can lose accuracy if the characterisations share few states. In the extreme case, if no states are shared, the distance value is always the same.

To overcome this issue, we suggest that the similarity between states should also be considered in the distance metric, besides the count of each state. This way, the distance will maintain its accuracy, regardless of the number of shared states. Further studies are required in order to assess the viability of this approach.



\subsection{Sampled Average State}

Regarding the \emph{sampled average state} technique, the most important factor to study is the influence of the parameter $W$. This parameter controls the length of the characterisation and how accurately it captures the temporal component of the robots behaviour. In the aggregation task, there was no significant difference among the treatments with different $W$ values ($p$-value $<$ 0.05). On the other hand, in the resource sharing task there is a trend in the results: the higher the $W$, the better the performance of the evolutionary process, regarding the quality of the solutions. The treatment with $W=50$ delivers significantly higher fitness scores than the treatment with $W=1$ ($p$-value $<$ 0.05).

The reason for the different impact of the $W$ value in different tasks is still not clear. Our hypothesis is that the difference is due to the degree of behaviour regularity  necessary to solve each task. The aggregation task can be solved using a regular pattern of behaviour, almost a reactive approach. As such, a low $W$ value might be sufficient to adequately characterise the behaviour of the swarm. On the other hand, the resource sharing task requires a more sequential behaviour, which involves first finding the charging station, and then a different behaviour for sharing it with the other robots. As of consequence, higher $W$ values might be preferred, as they allow the sequential component of the behaviour to be adequately captured. Further experiments, with different tasks, are required to confirm or reject this hypothesis.

\section{Conclusion}

We proposed two generic similarity measures for the domain of evolutionary swarm robotics, and used them to drive novelty search. The proposed measures rely on the principle that by analysing the microscopic behaviour of the robots of the swarm, it is possible to obtain a characterisation of the swarm behaviour as whole. The microscopic behaviour of each robot is exclusively based on the sensor and effector values of the robots, keeping the characterisation completely independent from the experimenter's domain knowledge.


The proposed similarity measures were tested in two distinct tasks, and compared with carefully crafted domain dependent similarity measures. The results showed that the performance obtained with the generic measures is just slightly inferior to the performance obtained with the domain dependent measures, regarding the quality of the evolved solutions. In each task, the highest scoring generic measures were not significantly worse than the domain dependent measure. Furthermore, the results show that the advantages of novelty search identified in previous work~\cite{gomes12a} hold with the generic measures: novelty search excelled at bootstrapping the evolutionary process, and was successful in circumventing deception.

In the comparison between the proposed generic similarity measures, we found that the \emph{sampled average state} achieved the best results in both tasks. However, from a general perspective, this measure is associated with a number of limitations: (i) the characterisations can become too long if there is a high number of sensors/effectors and a high value of $W$ is necessary; (ii) it is not applicable to tasks where simulations can have different lengths; and (iii) in tasks where the robots of the swarm are performing different sub-tasks at the same time, averaging the sensor-effector states of all robots can result in a meaningless characterisation. On the other hand, the \emph{combined state count} measure does not suffer from these limitations, despite the inferior performance verified in the two tasks presented in this paper. As such, we contend that the state count approach should not be discarded, and it should be further improved in future work. More experiments, with different tasks, are also needed in order to determine how well our results generalise, and clarify which measures are more suitable for each type of task.

The use of novelty search with generic behaviour similarity measures, in combination with traditional fitness-based evolution, opens interesting possibilities in the domain of evolutionary swarm robotics. First, it facilitates the use of straightforward fitness functions. There is no need to shape the fitness function in order to avoid local maxima, since novelty search circumvents that issue, without depending on additional information provided by the experimenter. It is a step towards evolving complex solutions with minimal intervention from the experimenter. Second, generic measures can potentially be used to unveil a true diversity of solutions based on self-organisation, with the evolved diversity not being conditioned by the experimenter.

\bibliographystyle{abbrv}
\bibliography{geccobib}

\begin{thebibliography}{10}

\bibitem{sahin05b}
E.~Bahge\c{c}i and E.~\c{S}ahin.
\newblock {Evolving aggregation behaviors for swarm robotic systems: A
  systematic case study}.
\newblock In {\em {Swarm Intelligence Symposium}}, pages 333--340. IEEE Press,
  2005.

\bibitem{baldassarre03}
G.~Baldassarre, S.~Nolfi, and D.~Parisi.
\newblock {Evolving Mobile Robots Able to Display Collective Behaviors}.
\newblock {\em Artificial Life}, 9(3):255--268, 2003.

\bibitem{baldassarre07}
G.~Baldassarre, V.~Trianni, M.~Bonani, F.~Mondada, M.~Dorigo, and S.~Nolfi.
\newblock {Self-Organized Coordinated Motion in Groups of Physically Connected
  Robots}.
\newblock {\em IEEE Transactions on Systems, Man, and Cybernetics},
  37(1):224--239, 2007.

\bibitem{sahin05a}
E.~\c{S}ahin.
\newblock {Swarm robotics: from sources of inspiration to domains of
  application}.
\newblock In {\em {2004 International Conf. on Swarm Robotics}}, volume 3342 of
  {\em {LNCS}}, pages 10--20. Springer-Verlag, 2005.

\bibitem{cuccu11b}
G.~Cuccu and F.~J. Gomez.
\newblock {When Novelty Is Not Enough}.
\newblock In {\em {European Conf. on the Applications of Evolutionary
  Computation}}, volume 6624 of {\em {LNCS}}, pages 234--243. Springer-Verlag,
  2011.

\bibitem{doncieux10}
S.~Doncieux and J.-B. Mouret.
\newblock {Behavioral diversity measures for Evolutionary Robotics}.
\newblock In {\em {IEEE Congress on Evolutionary Computation}}, pages 1--8.
  IEEE Press, 2010.

\bibitem{gomes12a}
J.~Gomes, P.~Urbano, and A.~L. Christensen.
\newblock {Introducing novelty search in evolutionary swarm robotics}.
\newblock In {\em {International Conf. on Swarm Intelligence}}, volume 7461 of
  {\em {LNCS}}, pages 85--96. Springer-Verlag, 2012.

\bibitem{gomes12b}
J.~Gomes, P.~Urbano, and A.~L. Christensen.
\newblock {Progressive Minimal Criteria Novelty Search}.
\newblock In {\em {Ibero-American Conf. on Artificial Int.}}, volume 7637 of
  {\em {LNAI}}, pages 281--290. Springer-Verlag, 2012.

\bibitem{gomez09}
F.~J. Gomez.
\newblock {Sustaining diversity using behavioral information distance}.
\newblock In {\em {Genetic and Evolutionary Computation Conf.}}, pages
  113--120. ACM Press, 2009.

\bibitem{hugues06}
L.~Hugues and N.~Bredeche.
\newblock {Simbad: An Autonomous Robot Simulation Package for Education and
  Research}.
\newblock In {\em {From Animals to Animats 9}}, volume 4095 of {\em {LNCS}},
  pages 831--842. Springer-Verlag, 2006.

\bibitem{krcah10}
P.~Krcah.
\newblock {Solving deceptive tasks in robot body-brain co-evolution by
  searching for behavioral novelty}.
\newblock In {\em {International Conf. on Intelligent Systems Design and
  Applications}}, pages 284--289. IEEE Press, 2010.

\bibitem{stan10b}
J.~Lehman and K.~O. Stanley.
\newblock {Efficiently evolving programs through the search for novelty}.
\newblock In {\em {Genetic and Evolutionary Computation Conf.}}, pages
  837--844. ACM Press, 2010.

\bibitem{stan10a}
J.~Lehman and K.~O. Stanley.
\newblock {Revising the evolutionary computation abstraction: minimal criteria
  novelty search}.
\newblock In {\em {Genetic and Evolutionary Computation Conf.}}, pages
  103--110. ACM Press, 2010.

\bibitem{stan11a}
J.~Lehman and K.~O. Stanley.
\newblock {Abandoning Objectives: Evolution Through the Search for Novelty
  Alone}.
\newblock {\em Evolutionary Computation}, 19(2):189--223, 2011.

\bibitem{mouret11}
J.-B. Mouret.
\newblock {Novelty-Based Multiobjectivization}.
\newblock In {\em {New Horizons in Evolutionary Robotics}}, volume 341 of {\em
  {Studies in Computational Intelligence}}, pages 139--154. Springer-Verlag,
  2011.

\bibitem{mouret09}
J.-B. Mouret and S.~Doncieux.
\newblock {Overcoming the bootstrap problem in evolutionary robotics using
  behavioral diversity}.
\newblock In {\em {IEEE Congress on Evolutionary Computation}}, pages
  1161--1168. IEEE Press, 2009.

\bibitem{mouret12}
J.-B. Mouret and S.~Doncieux.
\newblock {Encouraging Behavioral Diversity in Evolutionary Robotics: An
  Empirical Study}.
\newblock {\em Evolutionary Computation}, 20(1):91--133, 2012.

\bibitem{nelson09}
A.~L. Nelson, G.~J. Barlow, and L.~Doitsidis.
\newblock {Fitness functions in evolutionary robotics: A survey and analysis}.
\newblock {\em Robotics and Autonomous Systems}, 57(4):345--370, 2009.

\bibitem{stan02}
K.~O. Stanley and R.~Miikkulainen.
\newblock {Evolving Neural Network through Augmenting Topologies}.
\newblock {\em Evolutionary Computation}, 10(2):99--127, 2002.

\bibitem{trianni03}
V.~Trianni, R.~Gro{\ss}, T.~H. Labella, E.~\c{S}ahin, and M.~Dorigo.
\newblock {Evolving Aggregation Behaviors in a Swarm of Robots}.
\newblock In {\em {European Conf. on Artificial Intelligence}}, volume 2801 of
  {\em {LNCS}}, pages 865--874. Springer-Verlag, 2003.

\bibitem{trianni06a}
V.~Trianni, S.~Nolfi, and M.~Dorigo.
\newblock {Cooperative hole avoidance in a swarm-bot}.
\newblock {\em Robotics and Autonomous Systems}, 54(2):97--103, 2006.

\bibitem{whitley91}
L.~D. Whitley.
\newblock {Fundamental Principles of Deception in Genetic Search}.
\newblock In {\em {Foundations of Genetic Algorithms}}, pages 221--241. Morgan
  Kaufmann, 1991.

\bibitem{zaera96}
N.~Zaera, D.~Cliff, and J.~Bruten.
\newblock {({Not}) {E}volving Collective Behaviours in Synthetic Fish}.
\newblock In {\em {International Conf. on the Simulation of Adaptive
  Behavior}}, pages 635--644. MIT Press, 1996.

\end{thebibliography}

\end{document}